\documentclass[letterpaper, 10 pt, conference]{ieeeconf} 

\IEEEoverridecommandlockouts                        

\usepackage{graphics}
\usepackage{amsmath} 
\usepackage{amssymb}
\usepackage{cite}
\usepackage{tikz}
\usepackage{amsmath}
\usepackage{tabularx}
\usepackage{array}
\usepackage{booktabs}
\usepackage{bm}
\usepackage{algorithm}
\usepackage{algorithmic}
\usepackage{bbm}
\usepackage{adjustbox}
\usepackage{nicematrix}
\usepackage{graphicx}
\usepackage{caption}
\usepackage{wrapfig}
\usepackage{flushend}

\usepackage{tablefootnote}

\usepackage{hyperref}

\definecolor{green}{RGB}{11,155,13}

\DeclareMathOperator*{\argmin}{argmin}

\title{\LARGE \bf
Traverse the Non-Traversable: Estimating Traversability for\\ Wheeled Mobility on Vertically Challenging Terrain
}

\author{Chenhui Pan$^{*}$, Aniket Datar$^{*}$, Anuj Pokhrel, Matthew Choulas, Mohammad Nazeri, and Xuesu Xiao
\thanks{All authors are with the Department of Computer Science, George Mason University {\tt\scriptsize \{cpan7, adatar, apokhre, mnazerir, xiao\}@gmu.edu, matthewchoulas@gmail.com}}
\thanks{*Equally contributing authors}
}

\begin{document}
\maketitle
\thispagestyle{empty}
\pagestyle{empty}

\begin{abstract}
Most traversability estimation techniques divide off-road terrain into traversable (e.g., pavement, gravel, and grass) and non-traversable (e.g., boulders, vegetation, and ditches) regions and then inform subsequent planners to produce trajectories on the traversable part. 
However, recent research demonstrated that wheeled robots can traverse vertically challenging terrain (e.g., extremely rugged boulders comparable in size to the vehicles themselves), which unfortunately would be deemed as non-traversable by existing techniques. 
Motivated by such limitations, this work aims at identifying the traversable from the seemingly non-traversable, vertically challenging terrain based on past kinodynamic vehicle-terrain interactions in a data-driven manner. 
Our new Traverse the Non-Traversable (\textsc{tnt}) traversability estimator can efficiently guide a downstream sampling-based planner containing a high-precision 6-DoF kinodynamic model, which becomes deployable onboard a small-scale vehicle. 
Additionally, the estimated traversability can also be used as a costmap to plan global and local paths without sampling.
Our experiment results show that \textsc{tnt} can improve planning performance, efficiency, and stability  by {50}\%, {26.7}\%, and {9.2}\% respectively on a physical robot platform. 

\end{abstract}

\section{Introduction}

\label{sec::introduction}

Autonomous navigation in off-road environments is an exciting frontier in robotics research.
Its ever-growing applications in search and rescue, planetary exploration, mining, and agriculture warrant extensive research and development in off-road robot mobility.
The unpredictable nature of off-road terrain combined with the high risk of catastrophic failures presents significant challenges for mobile robots.
Navigating through \textit{vertically challenging} terrain with \textit{wheeled} robots~\cite{datar2024toward} in particular can cause robots to flip over, become airborne, or get stuck on the underlying terrain, especially if the obstacles are comparable in size to the robots themselves.

Most existing navigation systems classify off-road terrain into traversable and non-traversable spaces~\cite{castro2023traversability}.
Robots then employ path and motion planners to traverse on the traversable terrain, usually composed of free spaces on top of stable ground with minimal slope, to safely reach the goal. 
However, in extremely challenging  or time-critical missions, solely planning within traversable regions may not be possible or effective. Traversable terrain may not exist and limiting plans to only traversable regions can lead to overly conservative paths and delays in mission completion.
In such scenarios, more aggressive and risky maneuvers through so-called non-traversable terrain may be necessary to achieve timely results.

\begin{figure}
  \centering
  \includegraphics[width=\columnwidth]{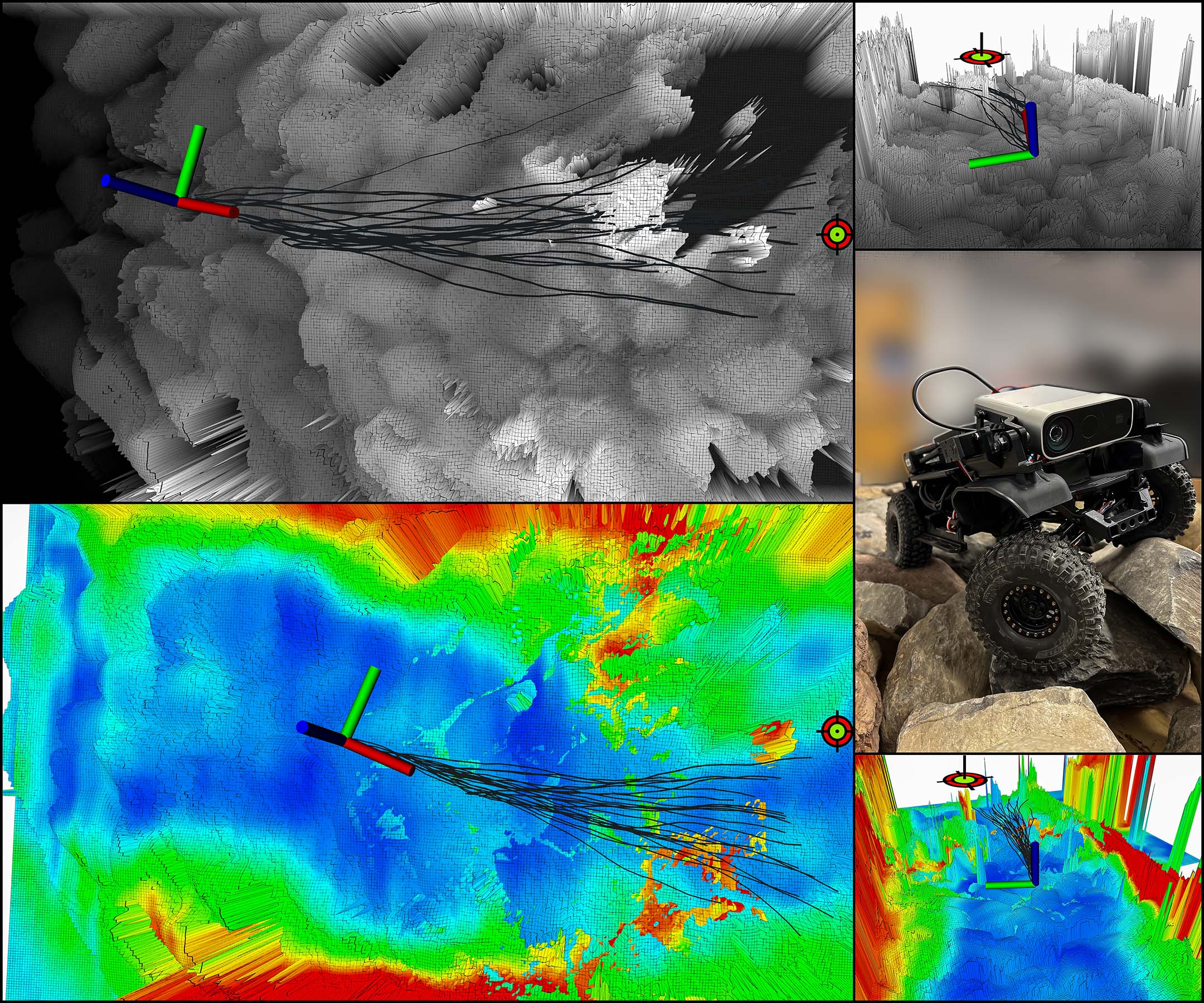}
  \caption{Guided by our \textsc{tnt} traversability estimator, a sampling-based motion planner with a high-precision kinodynamic model quickly converges to safe trajectories on vertically challenging terrain (bottom, colored traversability map), whereas without the \textsc{tnt} traversability map the planner gets stuck at exploring completely non-traversable area (top, white boulder on the black-white elevation map).}
  \label{fig::overview}
\end{figure}

Recent work on wheeled mobility has shown that even conventional wheeled vehicles have untapped potential to achieve impressive mobility on vertically challenging terrain~\cite{datar2024terrain, datar2023learning, datar2024toward}, with only minimal hardware requirements such as all-wheel drive, independent suspensions, and differential locking. 
This extended capability highlights that with the right combination of hardware and navigation strategies, even simple wheeled robots can overcome obstacles that were previously deemed non-traversable by state-of-the-art autonomous navigation systems.
Apart from end-to-end learning~\cite{datar2024toward}, another way to explore the possibility of traversing previously non-traversable terrain is through 6-DoF kinodynamic modeling to predict vehicle-terrain interaction and sampling-based planning to rollout potential future trajectories for evaluation~\cite{datar2023learning, datar2024terrain}. 
However, to deploy such methods robots must perform continuous and high-volume sampling across all terrain on any potential future paths.
This process involves numerous queries to a complex kinodynamic model to estimate traversability through a set of high-precision, multi-step 6-DoF trajectory rollouts. 
Such a combination is computationally demanding for resource-limited mobile robots.

In this work, we introduce Traverse the Non-Traversable (\textsc{tnt}), a terrain traversability estimator for previously non-traversable, vertically challenging terrain.
Based on a terrain elevation map, \textsc{tnt} generates a traversability map using previous vehicle-terrain interactions, including roll and pitch angles, discrepancy in vehicle command execution, and  uncertainty of kinodynamic model prediction. 
Instead of sampling the entire terrain elevation map and potentially getting stuck on a local optima, \textsc{tnt} filters out obviously non-traversable terrain in advance and allows a motion planner to only explore potentially traversable (while seemingly non-traversable) regions so as to quickly converge to a globally optimal plan. 
While \textsc{tnt} can guide sampling-based planners like \textsc{mppi}~\cite{williams2016aggressive} to improve sample efficiency, it can also be used as a costmap to plan both global and local paths through vertically challenging terrain without sampling.
Our physical experiments show that our \textsc{tnt} traversability estimation method can enable both a state-of-the-art sampling-based planner with a high-accuracy 6-DoF kinodynamic model and a search-based planner using a traversability cost function to traverse previously non-traversable, vertically challenging terrain. \textsc{tnt} achieves up to {50}\%, {26.7}\%, and {9.2}\% improvement in success rate, traversal time, and vehicle stability. 
\section{Related Work}
\label{sec::related_work}
Much of the work on off-road navigation originated from the DARPA Grand Challenge~\cite{seetharaman2006unmanned} and the LAGR~\cite{jackel2006darpa} program. Research and development in this field have inspired roboticists to push the boundaries of hardware~\cite{talia2023hound, datar2023learning}, perception~\cite{jiang2021rellis, procopio2009learning, viswanath2021offseg, wigness2019rugd}, planning~\cite{xiao2020appld, wang2021appli, wang2021apple, xu2021applr, xiao2022appl}, modeling~\cite{pokhrel2024cahsor, datar2024terrain, pagot2020real, karnan2022vi, xiao2021learning}, control~\cite{williams2016aggressive, xiao2021toward}, and learning~\cite{karnan2023sterling, kahn2021badgr, xiao2021toward, xiao2021agile, wang2021agile} in robotics. 
In this section, we briefly discuss related work to \textsc{tnt} in terms of traversability estimation and sampling-based planning.

\subsection{Terrain Traversability Estimation}
Beyond simple obstacle avoidance, accurate terrain traversability estimation is essential for safe and efficient navigation on off-road terrain. Numerous studies have explored various approaches to this problem, e.g., learning from vehicle-terrain interactions~\cite{wellhausen2019where, sathyamoorthy2022terrapn} and using a kinodynamic model~\cite{gasparino2022wayfast} to identify traversable paths and avoid obstacles. Recent works have leveraged both geometric~\cite{ruetz2024foresttrav} and semantic~\cite{schmid2022self, meng2023terrainnet, jung2024v} modalities for scene understanding and predicting environment elements~\cite{shaban2022semantic} beyond the perception range. Castro \emph{et al.},~\cite{castro2023does} estimated terrain traversability using Bird's Eye View (BEV) images and height maps, supervised by pseudo-ground truth cost derived from IMU data. The resulting costmap guides the robot towards traversable paths. While simpler approaches, such as classifying terrain as traversable or non-traversable~\cite{seo2023learning, frey2023fast}, have proven effective on flat off-road terrain, estimating traversability on vertically challenging terrain requires a more nuanced approach that extends beyond the simple binary classification. Furthermore, real-time decision-making is also crucial during deployment to ensure safe and efficient navigation. Sampling-based planners require continuous planning to facilitate convergence to optimal paths hence requiring quick traversability estimation. Roadrunner~\cite{frey2024roadrunner} achieves this by fusing sensor information with pre-trained image segmentation into a unified BEV representation.

Recent research has demonstrated the importance of accounting for both aleatoric uncertainty, which arises from partial observability~\cite{ewen2022these, cai2022risk}, and epistemic uncertainty, which stems from model distribution shift~\cite{seo2023scate, cai2024pietra}. While Endo \emph{et al.}~\cite{endo2023risk} focused solely on slip predictions for epistemic uncertainty, this limited scope proves insufficient for more complex, vertically challenging environments. EVORA~\cite{cai2024evora} presented a more comprehensive approach to traversability estimation by introducing a terrain traction model based on the ratio of commanded to realized velocities while also incorporating both aleatoric and epistemic uncertainties. This makes EVORA closely aligned with \textsc{tnt}, particularly in using commanded and realized velocity measurements and consideration of epistemic uncertainty to address terrain traversability challenges. 


\subsection{Sampling-based Planners}
Sampling-based planners have proven to be effective tools for finding optimal paths through complex terrain. While many of these planners operate successfully in 2D spaces, navigating challenging off-road environments often necessitates planners that can operate in $\mathbb{SE}(3)$~\cite{liu2018search} to generate agile trajectories in cluttered environments.

Search-based motion planners~\cite{datar2023learning, fox1997dynamic} use a deterministic approach to explore the configuration space. When the heuristic is well-defined, planners like A*~\cite{a*} and Dijkstra's are optimal. However, in an off-road environment, the high-dimensional configuration space makes achieving optimal performance with these planners very computationally expensive. This is where sampling-based planners~\cite{karaman2011sampling, gammell2014informed} have an advantage, as they explore the space flexibly by randomly sampling configurations. With the recent development of using a GPU~\cite{williams2017model, williams2016aggressive} to parallelize sampling, researchers have gravitated towards sampling in high volumes in the configuration space.

While sampling-based planners excel at efficiently finding solutions in complex off-road scenarios, they often suffer from a short planning horizon and lead to sub-optimal paths due to computation constraints. 
Guided planners~\cite{gammell2014informed} have emerged as a promising solution to this challenge. By leveraging information gained from the environment~\cite{sivaprakasam2021improving}, these planners predict what is beyond the planning horizon to prevent exploration of non-traversable regions and enable risk-aware~\cite{sharma2023ramp, beyer2024risk, tordesillas2021faster, xiao2020robot} planning in uncertain environments. 
Similarly, TNT can assist these planners by guiding their sampling region to only (potentially) traversable spaces.
\section{Approach}
\label{sec::approach}
Our hypothesis for the \textsc{tnt} estimator is that traversability is a simplified form of complex kinodynamic modeling. Prior works have shown that 6-DoF kinodynamic models in $\mathbb{SE}(3)$ are necessary for vertically challenging terrain~\cite{datar2024toward, datar2023learning, datar2024terrain}, which take into account current robot state, action, and underlying terrain patch to determine the next robot state. A full-scale 6-DoF model, while being precise, cannot be efficiently queried many times in a sampling-based motion planner to reveal the optimal solution through vertically challenging terrain. The key insight of \textsc{tnt} is that certain terrain patches, regardless of robot state and action, will induce undesired robot next state, i.e., being non-traversable. Identifying such terrain patches can limit the sampling space of sampling-based motion planners to only focus on potentially traversable areas in order to find the optimal motion plan to navigate through, or can serve as a costmap for search-based planners without sampling. 

\begin{figure}
  \centering
  \includegraphics[width=\columnwidth]{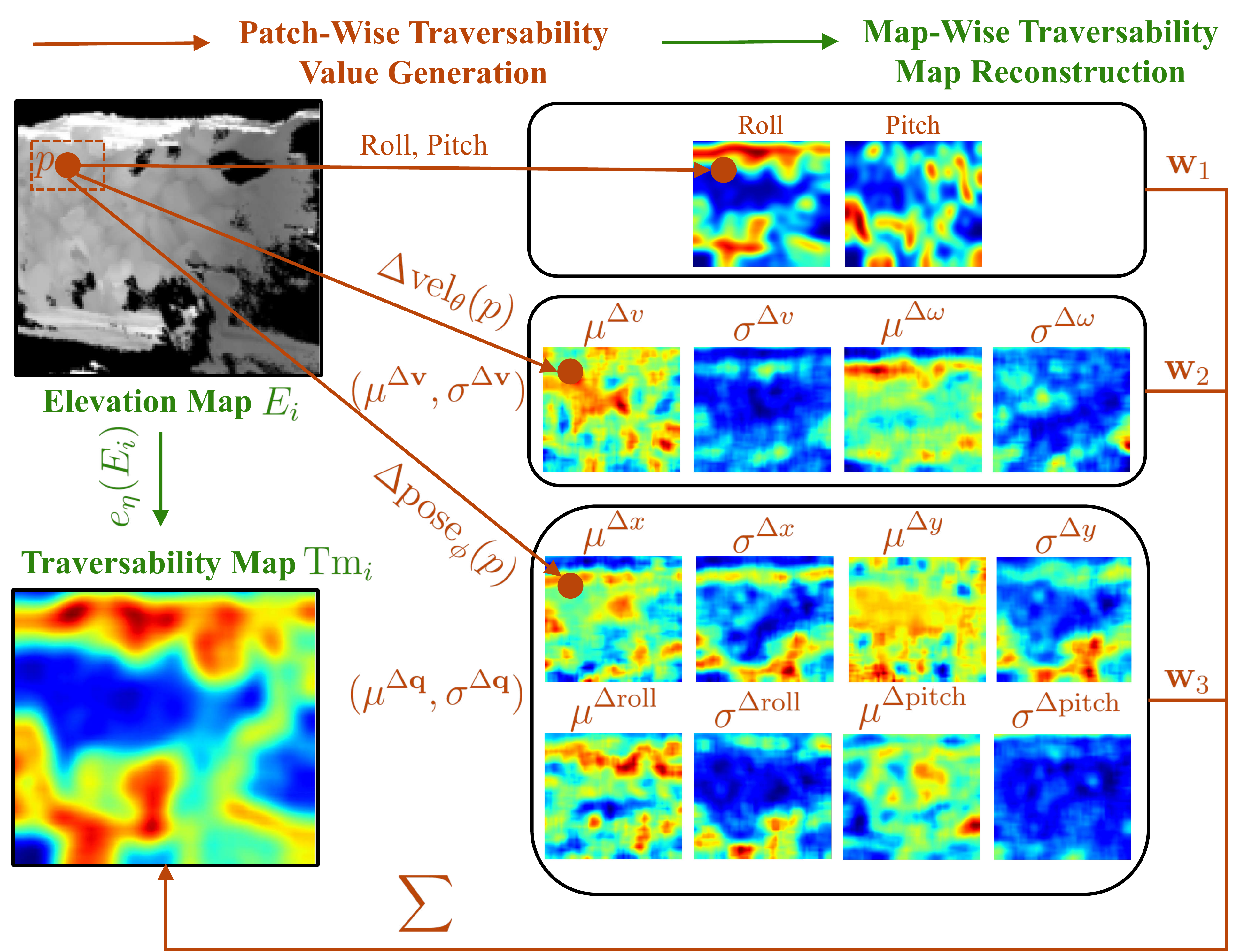}
  \caption{\textbf{\textsc{tnt} Overview.} Based on terrain patches $p$ on the elevation map $E_i$, three predictors produce roll and pitch angles , velocity uncertainty $ (\mu^{\Delta \mathbf{v}}, \sigma^{\Delta \mathbf{v}})$, and pose prediction uncertainty $(\mu^{\Delta \mathbf{q}}, \sigma^{\Delta \mathbf{q}})$, which are combined by $\mathbf{w}_1$, $\mathbf{w}_2$, and $\mathbf{w}_3$ to generate patch-wise traversability values; All traversability values form a traversability map $\text{Tm}_i$, which a map-wise traversability map estimator ($e_\eta(\cdot)$) learns to reconstruct based on terrain patch $E_i$. }
  \label{fig::tnt}
\end{figure}

Therefore, \textsc{tnt} aims to identify how (non-)traversable a terrain patch is regardless of robot state and action. We separate \textsc{tnt} into two stages: patch-wise traversability value generation and map-wise traversability map reconstruction. The former generates traversability value labels for the latter, which can be queried in real time onboard mobile robots (Fig.~\ref{fig::tnt}). 

\subsection{Patch-Wise Traversability Value Generation}
With the aforementioned simplification, \textsc{tnt} first estimates the traversability value of a terrain patch underneath the robot footprint regardless of robot state and action. The traversability value thus becomes a distribution over different robot states and actions. For vertically challenging terrain, we devise three intermediate metrics to comprise the final traversability value of a terrain patch $p$: (1) amplitude of roll and pitch angles, (2) velocity uncertainty in terms of the difference between commanded and actual vehicle velocities, and (3) pose prediction uncertainty in terms of the difference between predicted (based on a full 6-DoF model) and actual vehicle poses (from internal or external state estimation). 

\subsubsection{Roll and Pitch Angles}
\begin{wrapfigure}{r}{0.5\columnwidth}
\centering
\includegraphics[width=0.5\columnwidth]{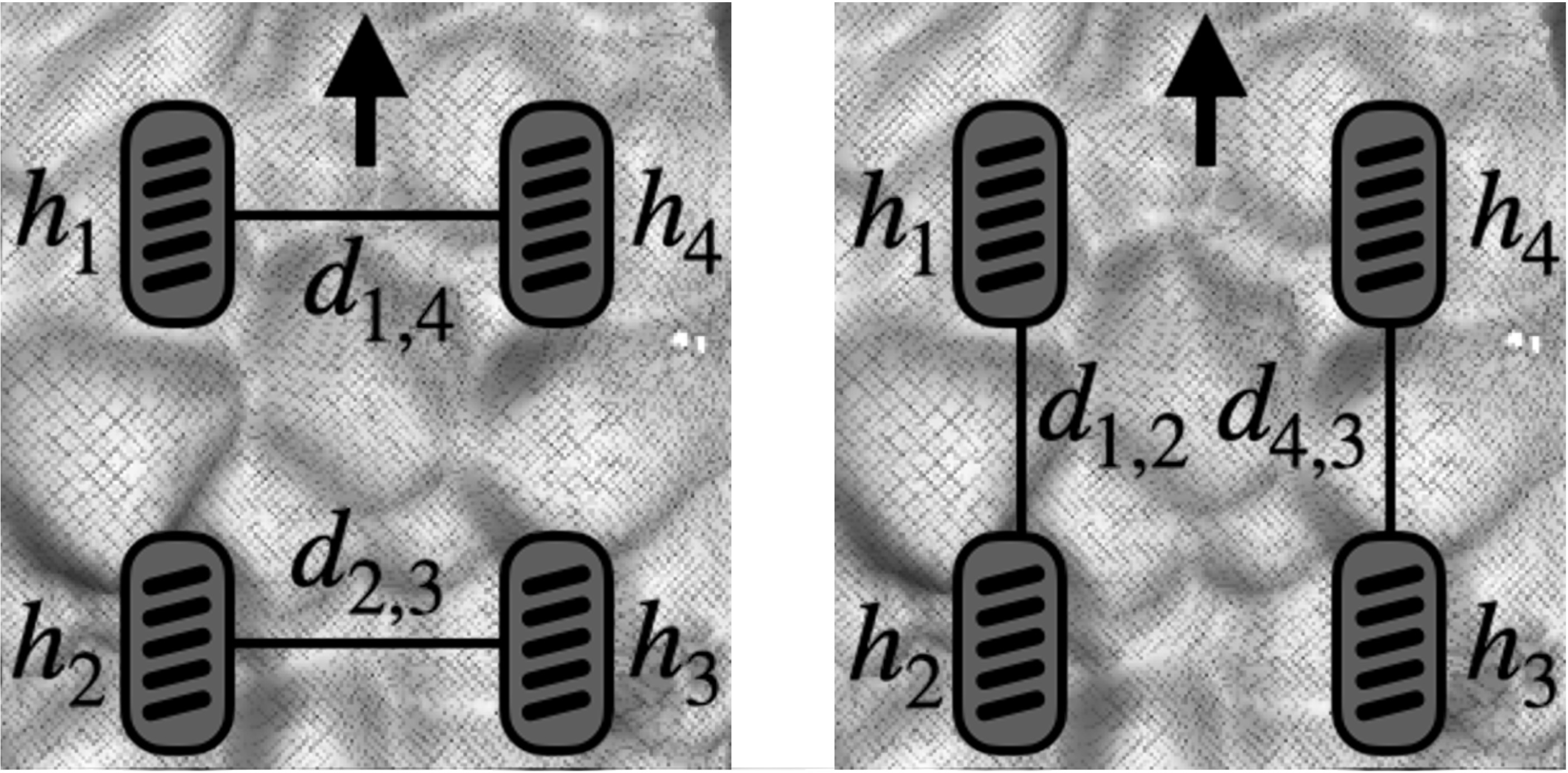}
\vspace{-1.5em}
\caption{Roll and Pitch Model.}
\vspace{-0.7em}
\label{fig::wheels}
\end{wrapfigure}
Two most frequent failure reasons on non-traversable vertically challenging terrain are vehicle rollover and getting-stuck due to excessive roll and pitch angles respectively. Therefore, the first component of the patch-wise traversability value is the amplitude of roll and pitch angles on a terrain patch based on a quasi-static physics-based model. 
The roll and pitch angles are estimated based on the elevation value $h_i$ of the terrain patch pixel under each vehicle wheel $i$ using trigonometry: 
\begin{equation}
 \text{roll}, \text{pitch} = \frac{1}{2}  \lVert \sum_{(i,j) \in \mathcal{I}_{\text{roll}, \text{pitch}}} \left(  \arctan \left( \frac{h_i-h_j}{d_{i,j}} \right) \right)\rVert, \label{eqn::wheels}
\end{equation}
where $d_{i,j}$ is the distance between the wheels $i,j$, and the subscript sets for roll and pitch are $\mathcal{I}_\text{roll} = \{(1,4), (2,3)\}$ (Fig.~\ref{fig::wheels} left) and $\mathcal{I}_\text{pitch} = \{(1,2), (4,3)\}$ (Fig.~\ref{fig::wheels} right), respectively. Notice that Eqn.~\eqref{eqn::wheels} only considers the terrain patch, not vehicle state and action, and the physics model is deterministic. 

\subsubsection{Velocity Uncertainty}
The intuition for the second component of the patch-wise traversability value is that less traversable terrain usually causes more significant uncertainty in terms of vehicle velocities. Particularly, we use the difference between commanded and actual vehicle velocities:
\begin{equation}
 \Delta \mathbf{v} = \mathbf{v}_\text{commanded} - \mathbf{v}_\text{actual},  \nonumber
\end{equation}
where $\mathbf{v} = (v, \omega)$ includes the linear and angular velocities of the vehicle. However, such a difference is dependent on the robot state and action. For example, if the vehicle is already traveling at a high speed, a steep slope won't cause too much difference between a high velocity command and the actual vehicle speed; However, a stationary vehicle starting on a steep slope with a high velocity command will cause a large difference. Therefore, an elevation map cannot uniquely determine the value of such differences, but rather a distribution over them. For example, extremely rugged terrain will cause a high probability of a large difference, while on a flat ground such a difference is mostly small. So the velocity uncertainty estimator is devised as:
\begin{equation}
 (\mu^{\Delta \mathbf{v}}, \sigma^{\Delta \mathbf{v}}) = \Delta\text{vel}_\theta(p), \label{eqn::vel}
\end{equation}
where $\mu^{\Delta \mathbf{v}}, \sigma^{\Delta \mathbf{v}}$ are the mean and standard deviation of the difference in linear ($\Delta v$) and angular ($\Delta \omega$) velocity, whereas the estimator $\Delta\text{vel}_\theta(\cdot)$ is parameterized by learnable parameters $\theta$. 

Given a dataset of ground truth $\mathcal{D}_{\Delta \mathbf{v}} = \{\Delta \mathbf{v}_i, p_i \}_{i=1}^N = \{(\Delta v_i, \Delta \omega_i), p_i \}_{i=1}^N$ computed by the difference between the outputs of a vehicle state estimator (e.g., internal Visual-Inertial Odometry or external GPS-RTK or motion capture system) and velocity commands, we represent $\Delta\text{vel}_\theta(\cdot)$ as a learnable neural network and learn the parameters $\theta$ using a negative log likelihood loss: 
\begin{equation}
 \mathcal{L}_{\mathcal{D}_{\Delta \mathbf{v}}} = \sum_{\substack{\{\Delta \mathbf{v}_i, p_i \} \\\in \mathcal{D}_{\Delta \mathbf{v}}}} \frac{1}{2} \left( \text{log}({\sigma_i^{\Delta \mathbf{v}}}^2(p_i))+\frac{(\mu_i^{\Delta \mathbf{v}}(p_i)-{\Delta \mathbf{v}}_i)^2}{{\sigma_i^{\Delta \mathbf{v}}}^2(p_i)} \right). 
 \label{eqn::vel_loss}
\end{equation}

\subsubsection{Pose Prediction Uncertainty}
The last component of the patch-wise traversability value is based on the intuition that it is more difficult to predict vehicle pose on less traversable terrain due to the more complex vehicle-terrain interactions. Therefore, we employ an existing kinodynamic model to predict the pose on a terrain patch $\mathbf{q}_\text{predicted}$ and compare it against the actual pose on that patch $\mathbf{q}_\text{actual}$: 
\begin{equation}
 \Delta \mathbf{q} = \mathbf{q}_\text{predicted} - \mathbf{q}_\text{actual}. \nonumber
\end{equation}
Pose $\mathbf{q}$ can include a 6-DoF vehicle state, i.e., $x$, $y$, $z$, roll, pitch, and yaw, and/or their higher-order derivatives. Since pose prediction using a forward kinodynamic model is iterative, the predicted $x$ and $y$ components are highly dependent on yaw, so including yaw is redundant in terms of uncertainty. Furthermore, the uncertainty in height $z$ does not provide significantly more insight compared to other pose components. Hence, in our implementation, we use the most informative components $x$, $y$, roll, and pitch (Fig.~\ref{fig::tnt}). Similar to Eqn.~\eqref{eqn::vel}, for pose prediction uncertainty, we employ another data-driven estimator: 
\begin{equation}
 (\mu^{\Delta \mathbf{q}}, \sigma^{\Delta \mathbf{q}}) = \Delta\text{pose}_\phi(p), \label{eqn::pose}
\end{equation}
parameterized by learnable parameters $\phi$. Another negative log likelihood loss, similar to Eqn.~\eqref{eqn::vel_loss} but with the $\Delta \mathbf{v}$ components replaced by $\Delta \mathbf{q}$, is used to learn $\phi$ in Eqn.~\eqref{eqn::pose}:
\begin{equation}
 \mathcal{L}_{\mathcal{D}_{\Delta \mathbf{q}}} = \sum_{\substack{\{\Delta \mathbf{q}_i, p_i \} \\\in \mathcal{D}_{\Delta \mathbf{q}}}} \frac{1}{2} \left( \text{log}({\sigma_i^{\Delta \mathbf{q}}}^2(p_i))+\frac{(\mu_i^{\Delta \mathbf{q}}(p_i)-{\Delta \mathbf{q}}_i)^2}{{\sigma_i^{\Delta \mathbf{q}}}^2(p_i)} \right).\nonumber 
\end{equation}
In our implementation, we use the Terrain-Attentive Learning (\textsc{tal}) model~\cite{datar2024terrain} with {25}-step prediction to produce $\mathbf{q}_\text{predicted}$. 

\subsubsection{Final Traversability Value}
To determine the final traversability value for the terrain patch $p$, we combine the roll and pitch prediction in Eqn.~\eqref{eqn::wheels}, the velocity uncertainty in Eqn.~\eqref{eqn::vel}, and pose prediction uncertainty in  Eqn.~\eqref{eqn::pose}: 
\begin{align} 
 & \text{Traversability}(p)= \nonumber \\
 &\mathbf{w}_1 \cdot [\text{roll}, \text{pitch}]^T + \mathbf{w}_2 \cdot [\mu^{\Delta \mathbf{v}}, \sigma^{\Delta \mathbf{v}}]^T + \mathbf{w}_3 \cdot [\mu^{\Delta \mathbf{q}}, \sigma^{\Delta \mathbf{q}}]^T,
 \label{eqn::traversability}
\end{align}
where $\mathbf{w}_1$, $\mathbf{w}_2$, and $\mathbf{w}_3$ are weight vectors to prioritize roll and pitch prediction, velocity uncertainty, and pose prediction uncertainty, along with their internal components, e.g., roll and pitch, linear and angular velocities, and 6-DoF pose dimensions.

\subsection{Map-Wise Traversability Map Reconstruction}
In principle, the patch-wise traversability value generation can be directly used during deployment, i.e., estimating the traversability of each terrain patch of interest, e.g., along the sampled vehicle trajectories to compute trajectory traversal costs. However, deploying all three models (Eqns.~\eqref{eqn::wheels}, \eqref{eqn::vel}, and \eqref{eqn::pose}) on each terrain patch located at and aligned with every vehicle state along hundreds or thousands of sampled trajectories is extremely computational intensive and therefore would defeat the purpose of traversability estimation (consider one could rather query the high-precision 6-DoF kinodynamic model for trajectory cost estimation~\cite{datar2024terrain}). Therefore, we use patch-wise traversability value generation to generate training data to learn a more efficient map-wise traversability map estimator. 
When moving through vertically challenging terrain, upcoming elevation maps in front of the robot are updated using onboard sensors (e.g., RGB-D camera or 3D LiDAR) and vehicle odometry at a low frequency (e.g., 2Hz). Our traversability map estimator takes as input updated elevation maps, and produces traversability values for each elevation pixel in the form of a traversability map. 

During training, we collect a variety of elevation maps $\{E_i\}_{i=1}^{M}$. On each elevation map $E_i$, we sample many terrain patches $p_j$ by varying the position indices, $m \sim \{1, 2, ..., H\}$ and $n \sim \{1, 2, ..., W\}$, and orientation angle, $\psi \sim \Psi$, where $H$ and $W$ are the height and width of $E_i$ and $\Psi$ is the set of candidate angles (e.g., $\Psi = \{-\frac{\pi}{2}, -\frac{\pi}{4}, 0, \frac{\pi}{4}, \frac{\pi}{2}\}$). We then acquire $p_i^{m, n, \psi} = g(E_i, m, n, \psi)$ with function $g(\cdot, \cdot, \cdot, \cdot)$ producing the patch centered at $(m, n)$ aligned with $\psi$ on $E_i$. For each $p_i^{m, n, \psi}$, we query the three estimators for roll and pitch (Eqn.~\eqref{eqn::wheels}), velocity uncertainty (Eqn.~\eqref{eqn::vel}), and pose prediction uncertainty (Eqn.~\eqref{eqn::pose}) in order to compute the traversability value $\text{Tr}_i^{m, n, \psi}$ for $p_i^{m, n, \psi}$ (Eqn.~\eqref{eqn::traversability}. For each pixel $(m, n)$ in $E_i$, we define the pixel value on the traversability map as:
\begin{equation}
    \text{Tm}_i^{m, n} = \frac{1}{|\Psi|} \sum_{\forall \psi \in \Psi}\text{Tr}_i^{m, n, \psi}, 
\end{equation}
i.e., the traversability map pixel value $\text{Tm}_i^{m, n}$ at position $(m, n)$ is averaged over all traversability values $\text{Tr}_i^{m, n, \psi}$ at all possible vehicle orientations $\forall \psi \in \Psi$. All individual $\text{Tm}_i^{m, n}, \forall m, n$ compose the full traversability map $\text{Tm}_i$. For each elevation map $E_i$ and traversability map $\text{Tm}_i$, we train a traversability map encoder $e_\eta(\cdot)$ with learnable parameters $\eta$, in order to minimize a traversability map reconstruction loss: 
\begin{equation}
    \eta^* = \argmin_{\eta} \sum_{i=1}^{M} \lVert e_\eta(E_i)-\text{Tm}_i\rVert. 
\end{equation}
$e_\eta(\cdot)$ will be used online to produce $\text{Tm}_i$ when a new elevation map $E_i$ is available to guide a sampling-based planner or serve as a costmap for a search-based planner.

\section{Experiments}
\label{sec::experiments}
In this section, we present our experiment results of the \textsc{tnt} traversability estimator. We visualize the \textsc{tnt} traversability map and show that search-based planners like A* can be used to find the most traversable path. We also use \textsc{tnt} in a sampling-based motion planner with a high-precision 6-DoF kinodynamic model and demonstrate improved sample efficiency and navigation performance.   

\subsection{Implementations}
\subsubsection{Robot, Testbed, and Data}
We implement \textsc{tnt} on the Verti-4-Wheeler (V4W), an open-source, 1/10th-scale robotic platform~\cite{datar2024toward}. V4W is equipped with advanced mobility features, including a low-high gear switch and lockable front and rear differentials, which enhance its performance on vertically challenging terrain. The perception system comprises a Microsoft Azure Kinect RGB-D camera, while an OptiTrack motion capture system provides odometry data during training data collection only, but not for experiments. Real-time elevation mapping is facilitated by an open-source tool~\cite{miki2022elevation}, which processes depth input from the Azure Kinect camera. Computational tasks are managed by an onboard NVIDIA Jetson Orin NX computer.

To facilitate experimentation, we construct a scaled testing environment. The testbed consists of hundreds of rocks and boulders, measuring 3.1 m$\times$1.3 m with a maximum height of 0.6 m. This environment is highly reconfigurable, allowing for a wide range of experimental scenarios.

The dataset collected for traversability estimation encompasses human-teleoperated vehicle controls (throttle and steering commands), elevation maps derived from depth images, and odometry data from the motion capture system for vehicle state estimation. To ensure comprehensive coverage, the dataset includes a diverse range of 6-DoF robot states, with particular emphasis on capturing maximum roll and pitch conditions. Additionally, the robot is operated at varying speeds within the same rock configuration to gather data for velocity uncertainty estimation. The resulting dataset consists of 70,816 individual elevation maps and 139 minutes of vehicle teleoperation on the rock testbed. 

\subsubsection{Architectures}
The prediction models for velocity uncertainty and pose uncertainty consist of a combination of a 7-layer Convolutional Neural Network (CNN) followed by three fully connected layers with the output dimensions of four (mean and standard deviation of linear and angular velocity, $v$ and $\omega$) and eight (mean and standard deviation of $x$, $y$, roll, and pitch) respectively (Fig.~\ref{fig::tnt}). 

The map-wise traversability map estimator consists of a 4-layer CNN with residual connections to encode the elevation map. The encoding is passed through a 3-layer convolution transpose followed by upsampling to construct a 14-channel map of size 320$\times$260, including two channels for roll and pitch, four for velocity uncertainty, and eight for pose uncertainty. This traversability map estimator is the final model used during deployment to generate the traversability map in real-time onboard the robot. The final traversability map is generated by calculating a weighted sum of all 14 channels. These weights can be fine-tuned for different scenarios and also dynamically adjusted based on the current robot state. 

\subsection{Traversability Map Visualization and A* Planning}
\begin{figure}
  \centering
  \includegraphics[width=\columnwidth,]{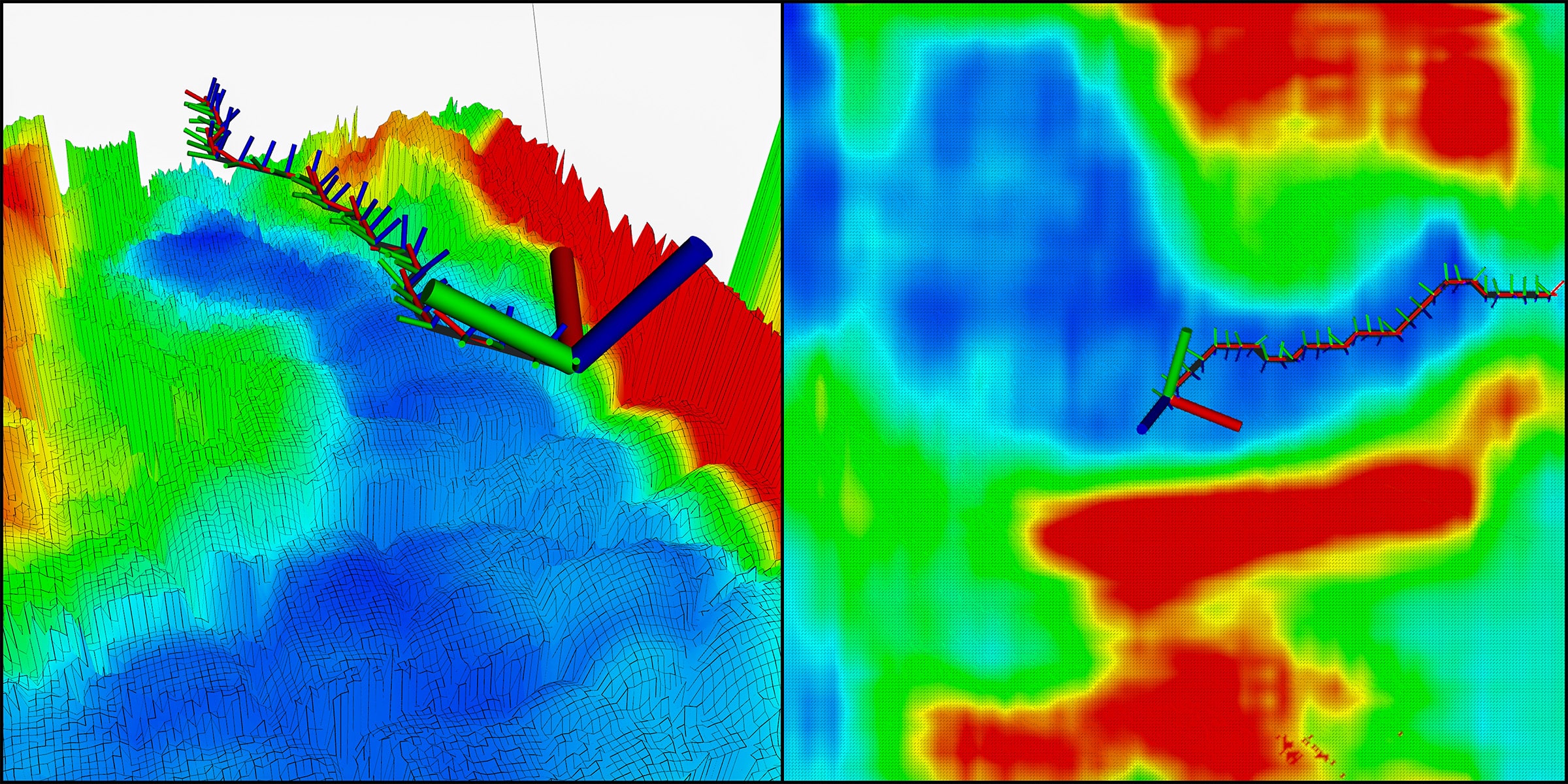}
  \caption{\textbf{Traversability Map Generated by \textsc{tnt}}. The color gradient represents traversability, with blue areas indicating easily traversable terrain and red areas signifying challenging or non-traversable regions. The overlaid path represents the optimal route calculated by the A* algorithm.}
  \label{fig::visualization}
\end{figure}

We showcase an example of the traversability map produced by our \textsc{tnt} traversability estimator in Fig.~\ref{fig::visualization}. Despite the vertical challenges caused by the rocks and boulders, \textsc{tnt} is able to effectively discern the obviously non-traversable terrain patches from the potentially traversable ones, providing crucial terrain information to downstream mobility tasks, like path planning and motion control. Using the traversability map, we demonstrate that simple search-based planners, such as A*,  are able to plan the most traversable path through vertically challenging terrain, without considering vehicle dynamics. 

To be specific, we first down-sample the final traversability map to a 31$\times$25 grid using a convolution filter to  average the pixel values. Then, the A* algorithm can efficiently compute the optimal path from the robot position to the goal location, minimizing both the traversability cost and Euclidean distance.  The A* path can be converted to desired robot positions to be tracked by downstream local planners or controllers, such as the DWA planner~\cite{fox1997dynamic} or a PID controller. 

\subsection{Physical Experiments}
We integrate \textsc{tnt} with a sampling-based motion planner, \textsc{mppi}~\cite{williams2017model}, and deploy it on a physical V4W robot. To address vertically challenging terrain, the \textsc{mppi} planner employs a state-of-the-art, high precision 6-DoF kinodynamic model, \textsc{tal}~\cite{datar2024terrain}, to minimize a cost function considering vehicle trajectories in $\mathbb{SE}(3)$. 
Previous work~\cite{datar2024terrain} demonstrated the computational difficulty in rolling out a large amount of 6-DoF trajectories  with high variance using \textsc{tal} to cover a large portion of the state space, which potentially includes the globally optimal trajectory, thereby compromising real-time planning and navigation performance. In fact, it has been shown that despite its higher kinodynamics accuracy, such a difficulty limits \textsc{tal}'s performance to that of a simplified 6-DoF model decomposed into a planar Ackerman, roll and pitch prediction, and elevation value model, i.e., the \textsc{wmvct} model and planner~\cite{datar2023learning}. 
Considering that such a limitation partially motivates \textsc{tnt} (Fig.~\ref{fig::overview}), we showcase the importance of efficiently biasing the sampling distribution of \textsc{mppi} using \textsc{tnt} to more efficiently utilize the high-precision, but also high-computation \textsc{tal} model. 

Specifically, we denote the \textsc{mppi} planner with the \textsc{tal} model as \textsc{tal}, whereas \textsc{tnt} indicates that they are augmented by \textsc{tnt}. We also compare with the \textsc{wmvct} model and planner, denoted as \textsc{wmvct}. Table \ref{tab::results} shows the experiment results on a randomly created test course on the vertically challenging testbed (Fig.~\ref{fig::tnt_experiments} left). 

As shown in Table \ref{tab::results} , \textsc{tnt} achieves better success rate and traversal time compared to \textsc{tal} and \textsc{wmvct}. This is because \textsc{mppi} without \textsc{tnt} is only guided by the accurate 6-DoF vehicle trajectories produced by the slow  \textsc{tal} model, which must rely on a long horizon in order to rollout until reaching certain terrain patch to determine its cost based on $\mathbb{SE}(3)$ vehicle state. Thus, the requirement on long horizon limits the sample number and variance to a low value due to onboard computation limitation. The robot then lacks efficient anticipation of all possibilities of future paths to explore traversable terrain while preventing completely non-traversable areas. 
On the other hand, due to the rich traversability information provided by \textsc{tnt}, \textsc{mppi} can plan with a shorter horizon but still anticipate what is coming up ahead, since all such information has already been efficiently distilled into our traversability map during training. Therefore, \textsc{tnt} helps \textsc{mppi} to  converge faster to regions that are more traversable. 
Guided by the \textsc{tnt} traversability map, \textsc{mppi} can rollout a smaller amount of shorter trajectories with a faster convergence time, but still quickly discover the most promising future vehicle trajectory to traverse through the vertically challenging terrain. 
Similar to the results reported by prior work~\cite{datar2024terrain}, \textsc{wmvct} performs similarly or better compared to \textsc{tal} with a less accurate decomposed 6-DoF kinodynamic model, thanks to the efficient model query time. 

We also report the changes in roll and pitch angles, as well as in throttle and steering commands resulted by the three methods. \textsc{tnt} achieves the lowest average and variance across all four metrics, demonstrating a very stable navigation behavior. In most cases, \textsc{tal} is less stable than \textsc{wmvct}, except change in roll. The smaller change values in all these four metrics achieved by \textsc{tnt} indicate that the \textsc{mppi} planner guided by \textsc{tnt} is able to efficiently find traversable paths, without wasting computation and movement effort on exploring undesired, less traversable areas. 


\begin{table}
\caption{\textbf{Experiment Results of \textsc{tnt}, \textsc{tal} and \textsc{wmvct}:} success rate, mean traversal time (of successful trials), mean absolute roll and pitch angles, and mean changes in roll, pitch, throttle, and steering.}
\centering
\resizebox{\columnwidth}{!}{%
\small
\setlength{\tabcolsep}{4pt}
\begin{tabular}{ccccccccc}
\toprule
                    & {\textsc{TNT}} & {\textsc{TAL}} & {\textsc{wmvct}} \\
\midrule
{Success Rate $\uparrow$}      & {\textbf{9/10}} & {6/10} & {8/10}\\
{Traversal Time $\downarrow$}      & {\textbf{17.6s}$\pm$2.64s} & {24.0s}$\pm$9.88s & {21.1s}$\pm$24.90s\\
{Absolute Roll $\downarrow$}      & \textbf{{6.6}\textdegree$\pm$7.0\textdegree} & {9.4\textdegree$\pm$24.6\textdegree} & {8.3\textdegree$\pm$11.9\textdegree}\\
{Absolute Pitch $\downarrow$}      & {8.2\textdegree$\pm$7.4}\textdegree & \textbf{{6.9}\textdegree$\pm$10.9\textdegree} & {8.8\textdegree$\pm$10.7\textdegree}\\
{$\Delta$ Roll $\downarrow$}      & {\textbf{0.51\textdegree$\pm$1.03\textdegree}} & {0.63\textdegree$\pm$6.07\textdegree} & {0.97\textdegree$\pm$1.6\textdegree}\\
{$\Delta$ Pitch $\downarrow$}      & {\textbf{0.53\textdegree$\pm$0.68\textdegree}} & {0.080\textdegree$\pm$1.3\textdegree} & {0.57\textdegree$\pm$1.08\textdegree}\\
{$\Delta$ Throttle $\downarrow$}      & {\textbf{0.042$\pm$0.15}} & {0.065}$\pm$0.34 & {0.053$\pm$0.31}\\
{$\Delta$ Steering $\downarrow$}      & {\textbf{0.071$\pm$0.24}} & {0.153$\pm$0.38} & {0.087$\pm$0.23}\\
\bottomrule
\end{tabular}%
}
\label{tab::results}
\end{table}

\subsection{Outdoor Demonstration}
\begin{figure}
\centering
\includegraphics[width=\columnwidth]{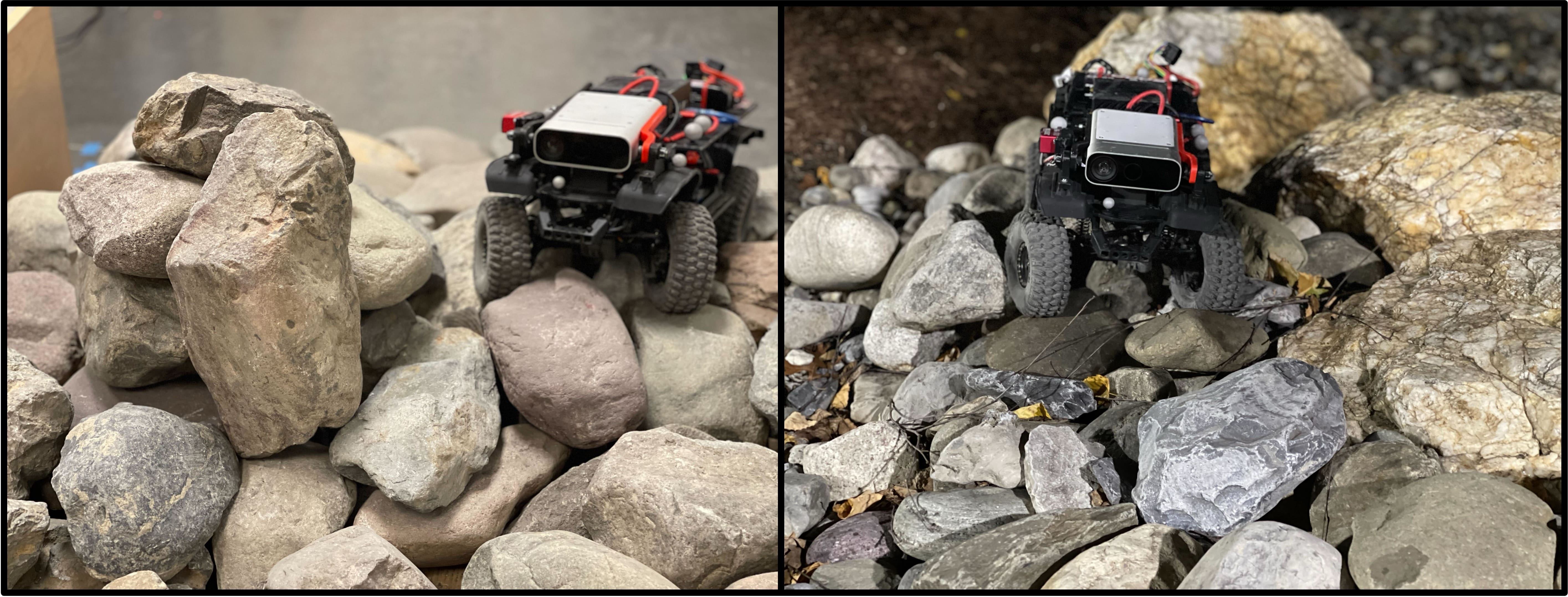}
\caption{\textsc{tnt} Indoor Experiments and Outdoor Demonstration.}
\label{fig::tnt_experiments}
\end{figure}

In addition to the indoor physical experiments on the vertically challenging testbed, we also demonstrate that \textsc{tnt} can be deployed in an outdoor off-road environment. The natural outdoor off-road environment is filled with pebbles, rocks, and boulders of a wide range of sizes. Grass, mulch, and gravel are also present next to the rocks. \textsc{tnt} is still able to produce accurate traversability maps to guide subsequent \textsc{mppi} planner to traverse through previously non-traversable, vertically challenging terrain (Fig.~\ref{fig::tnt_experiments} right).

\section{Conclusions}
\label{sec::conclusions}

We present Traverse the Non-Traversable (\textsc{tnt}), a traversability estimation framework that addresses the limitations of traditional traversability estimation techniques that struggle to identify traversable areas on vertically challenging terrain. The \textsc{tnt} traversability estimator, leveraging data-driven insights from past kinodynamic vehicle-terrain interactions and a physics-based model, enables robots to navigate areas that are deemed non-traversable by traditional traversability estimators. By integrating \textsc{tnt} with a high-precision kinodynamic model and a sampling-based motion planner, or by utilizing it as a costmap for path planning, significant improvements in planning performance, efficiency, and stability have been demonstrated on a physical robot platform. This work paves the way for robots to traverse previously inaccessible environments, expanding their operational capabilities across various domains.

Despite \textsc{tnt}'s efficacy in estimating terrain traversability from a purely geometric perspective, i.e., from 2.5D elevation maps, one interesting future direction is to incorporate semantics into the traversability estimator. Grass, mulch, mud, and gravel will exhibit different traversability, although they may look similarly to rocks and boulders in a geometric sense. Terrain granularity and deformability also need to be considered to comprehensively evaluate traversability. 

\newpage
\bibliographystyle{IEEEtran}
\bibliography{IEEEabrv,references}
\end{document}